\documentclass[conference]{IEEEtran}
\IEEEoverridecommandlockouts
\usepackage{cite}
\usepackage{amsmath,amssymb,amsfonts}
\usepackage{algorithmic}
\usepackage{graphicx}
\usepackage{textcomp}
\usepackage{xcolor}
\def\BibTeX{{\rm B\kern-.05em{\sc i\kern-.025em b}\kern-.08em
    T\kern-.1667em\lower.7ex\hbox{E}\kern-.125emX}}
\begin{document}

\title{Assessing the Potential of Masked Autoencoder Foundation Models in Predicting Downhole Metrics from Surface Drilling Data}

\author{
    \IEEEauthorblockN{1\textsuperscript{st} Aleksander Berezowski}
    \IEEEauthorblockA{\textit{Schulich School of Engineering} \\
    \textit{Department of Electrical and Software Engineering} \\
    \textit{University of Calgary}\\
    Calgary, Canada \\
    aleksander.berezowsk@ucalgary.ca}
\and
    \IEEEauthorblockN{2\textsuperscript{nd} Hassan Hassanzadeh}
    \IEEEauthorblockA{\textit{Schulich School of Engineering} \\
    \textit{Department of Chemical and Petroleum Engineering} \\
    \textit{University of Calgary}\\
    Calgary, Canada \\
    hhassanz@ucalgary.ca}
\and
    \IEEEauthorblockN{3\textsuperscript{rd} Gouri Ginde}
    \IEEEauthorblockA{\textit{Schulich School of Engineering} \\
    \textit{Department of Electrical and Software Engineering} \\
    \textit{University of Calgary}\\
    Calgary, Canada \\
    gouri.ginde@ucalgary.ca}
}

\maketitle

\begin{abstract}
Oil and gas drilling operations generate extensive time-series data from surface sensors, yet accurate real-time prediction of critical downhole metrics remains challenging due to the scarcity of labelled downhole measurements. This systematic mapping study reviews thirteen papers published between 2015 and 2025 to assess the potential of Masked Autoencoder Foundation Models (MAEFMs) for predicting downhole metrics from surface drilling data. The review identifies eight commonly collected surface metrics and seven target downhole metrics. Current approaches predominantly employ neural network architectures such as artificial neural networks (ANNs) and long short-term memory (LSTM) networks, yet no studies have explored MAEFMs despite their demonstrated effectiveness in time-series modeling. MAEFMs offer distinct advantages through self-supervised pre-training on abundant unlabeled data, enabling multi-task prediction and improved generalization across wells. This research establishes that MAEFMs represent a technically feasible but unexplored opportunity for drilling analytics, recommending future empirical validation of their performance against existing models and exploration of their broader applicability in oil and gas operations.
\end{abstract}


\section{Introduction}
Oil and gas wellbore drilling is a highly data-intensive process that requires continuous monitoring and 
control of numerous physical parameters to ensure operational efficiency. During drilling operations, 
vast volumes of time-series data are generated from surface sensors, including weight on bit (WOB), standpipe 
pressure (SPP), torque (T), and rotational speed (RPM) \cite{pr13051472}. These surface measurements are used to 
estimate downhole conditions that are difficult or costly to measure in real time, such as bottom-hole 
pressure (BHP) \cite{ZhangRui2023ANHT}, equivalent circulating density (ECD) \cite{10.56952/ARMA-2023-0722}, and 
downhole string vibrations \cite{SaadeldinRamy2022IMfP, SaadeldinRamy2023Ddvt}. Accurate prediction of these downhole 
metrics is critical for optimizing drilling performance.

In recent years, machine learning (ML) has emerged as a promising approach for inferring downhole metrics solely from surface measurements \cite{Tariq_Aljawad_Hasan_Murtaza_Mohammed_El-Husseiny_Alarifi_Mahmoud_Abdulraheem_2021}. A variety of ML models have been developed to map multiple surface features to single downhole targets \cite{EkechukwuGerald2024Empo, GamalHany2021MLMf, AbdelgawadKhaledZ.2019Nate, ZhaoWanchun2025Tdme, ZhangCheng-Kai2023Bhpp, SaadeldinRamy2022IMfP, SaadeldinRamy2023Ddvt, ZhangRui2023ANHT, ZhouYang2021Aohp, CAOWanpeng2025Dlat, ZhaoJie2017MLTD, EncinasMauroA.2022Ddcf, HegdeChiranth2015UTBa}. While these approaches have shown utility, they often rely on large labeled datasets for each target metric, struggle to generalize across wells, and are limited in their ability to support multiple predictive tasks simultaneously.

Masked Autoencoder Foundation Models (MAEFMs), first introduced by He et al. 
\cite{He_Chen_Xie_Li_Dollar_Girshick_2022}, constitute a fundamentally new class of ML models with capabilities 
that directly address these limitations. Building on the concept of latent-space representations originally 
proposed by Rumelhart et al. \cite{Rumelhart_Hinton_Williams_1986}, MAEFMs employ self-supervised pre-training 
to learn generalizable features from large unlabeled datasets and can subsequently be fine-tuned on relatively 
small labeled datasets for multiple downstream tasks. This paradigm is particularly well-suited to the oil and 
gas sector, where drilling operations generate abundant surface sensor data but labeled downhole measurements 
are scarce and expensive. MAEFMs have demonstrated state-of-the-art performance in both computer vision and 
time-series forecasting \cite{Mienye_Swart_2025}, yet their application to downhole metric prediction remains 
entirely unexplored. The potential to leverage large-scale drilling datasets to learn transferable latent 
representations, enable multi-task predictions, and reduce dependence on labeled downhole data represents a 
significant opportunity for research and innovation in the sector.

Given this context, it is important to evaluate the suitability of MAEFMs for oil and gas drilling applications, 
review existing ML approaches for downhole metric prediction, and identify performance limitations and knowledge 
gaps. This study aims to investigate the feasibility of MAEFMs for predicting critical downhole metrics, outline 
directions for future research, and highlight where this emerging technology could offer significant advantages 
in operational efficiency, safety, and cost reduction.

\section{Research Questions}
The research study is guided by four research questions, each with one or more subquestions, designed to 
analyze the application of MAEFMs in oil and gas drilling. 
\\

\textbf{Q1: What measurements are important to oil and gas drillers?}
\begin{itemize}
\item Q1.1: What surface metrics are collected?
\item Q1.2: What downhole metrics are important to predict?
\end{itemize}

Drilling is data-intensive, with numerous metrics collected at the surface and a desire to predict 
many downhole metrics. This question identifies what metrics can be used as inputs to ML models 
(i.e., metrics collected at the surface) and what metrics are the outputs of ML models (i.e., 
downhole metrics that are predicted), allowing for greater clarity of the scope of ML in oil 
and gas drilling.
\\

\textbf{Q2: What methods are used to predict downhole metrics?}
\begin{itemize}
\item Q2.1: What methods are currently used to predict downhole metrics from surface drilling data?
\end{itemize}

This question establishes how ML is currently used to predict downhole metrics, providing background context on current ML applications.
\\

\textbf{Q3: Can MAEFMs be used for oil and gas drilling?}
\begin{itemize}
\item Q3.1: What are MAEFMs?
\item Q3.2: Are MAEFMs suitable for time-series data such as surface drilling metrics?
\item Q3.3: Are MAEFMs used in oil and gas drilling currently?
\end{itemize}

This question evaluates the technical feasibility and novelty of MAEFMs in oil and gas drilling. This ensures that MAEFMs are suitable for oil and gas and are able to have an impact on the field.  
\\

\textbf{Q4: What should the research direction for MAEFMs in the context of oil and gas drilling 
be?}
\begin{itemize}
\item Q4.1: What areas could MAEFMs be applied to in oil and gas drilling?
\end{itemize}

This question focuses on defining where MAEFMs could have the most impact, and where research 
should focus. This helps define the best opportunity for MAEFM application in oil and gas 
drilling.

\section{Data Collection Methodology}
The data collection methodology is divided into three sections: paper search, inclusion and 
exclusion criteria, and data extraction. The methodology presented is consistent with 
methodologies presented by Peterson et al. \cite{Systematic2008}, making it a version of a 
systematic mapping study.

\subsection{Paper Search}

A search query was crafted to find papers that used time-series surface drilling data to predict 
time-series downhole metrics using ML. The search query was composed of four parts that were combined with AND operators. The first part ensured domain relevancy, ensuring the search targeted research in the 
drilling space. The second part narrowed the search to research focused on downhole metrics. The 
third part further narrowed the search to prediction, ensuring research was focused on predicting 
downhole metrics. The final part of the search ensured ML was being used to predict the downhole drilling metrics, as opposed to rules based algorithms. This resulted in the following search query:
\\
\begin{quote}
\texttt{("drilling data" OR "surface drilling measurements" OR "surface drilling parameters")}

\textbf{AND}

\texttt{("downhole metrics" OR "downhole parameters" OR "bottom hole pressure" OR}

\texttt{"equivalent circulating density" OR "downhole vibrations" OR "torque and drag")}

\textbf{AND}

\texttt{("prediction" OR "estimation" OR "forecasting" OR "time series")}

\textbf{AND}

\texttt{("machine learning" OR "neural network" OR "deep learning" OR "regression" OR}

\texttt{"artificial intelligence")}
\end{quote}
Additionally, the following search criteria were applied:
\begin{itemize}
\item Language: Published in English
\item Date Range: Published in or after January 2015 until November 2025
\item Article Type: Article or conference proceedings
\item Availability: Full text available online
\end{itemize}

The search was conducted across multiple databases, and the following databases returned at least 
one relevant result:
\begin{itemize}
    \item ScienceDirect
    \item OnePetro
    \item The American Society of Mechanical Engineers Digital Collection
    \item Directory of Open Access Journals
    \item National Library of Medicine
\end{itemize}

The search string and search criteria resulted in 20 results. 

\subsection{Inclusion and Exclusion Criteria}
The retrieved papers were evaluated through a two-stage screening process consisting of a title 
and abstract review followed by a full-text review. The inclusion and exclusion criteria outlined 
in Table~\ref{tab:criteria} were applied at both stages. Papers that did not meet the inclusion 
criteria or that met any of the exclusion criteria were excluded from the final selection.

\begin{table}[htbp]
\caption{Inclusion and Exclusion Criteria}
\centering
\begin{tabular}{|p{0.45\linewidth}|p{0.45\linewidth}|}
\hline
\textbf{Inclusion Criteria} & \textbf{Exclusion Criteria} \\
\hline
\begin{itemize}
    \item Used surface metrics as inputs to an ML model.
    \item Predicted one or more downhole metrics as an output for an ML model.
\end{itemize}
&
\begin{itemize}
    \item Focused on non-drilling applications like reservoir modeling or seismic studies.
    \item Was a duplicate study.
\end{itemize} \\
\hline
\end{tabular}
\label{tab:criteria}
\end{table}

After applying the criteria, there were 13 papers, as shown in Table~\ref{tab:reviewed-papers}.

\begin{table*}[htbp]
\caption{Summary of Reviewed Papers}
\centering
\begin{tabular}{|c|p{3.5cm}|p{2.8cm}|p{3.5cm}|c|c|}
\hline
\textbf{Paper} & \textbf{Title} & \textbf{First and Last Authors} & \textbf{Author Affiliations} & \textbf{Citations} & \textbf{Year} \\
\hline
1 & Explainable machine-learning-based prediction of equivalent circulating density using surface-based drilling data \cite{EkechukwuGerald2024Empo} & Gerald Ekechukwu; Abayomi Adejumo & Louisiana State University; Oriental Energy Resources Limited & 7 & 2024 \\
\hline
2 & Machine Learning Models for Equivalent Circulating Density Prediction from Drilling Data \cite{GamalHany2021MLMf} & Hany Gamal; Salaheldin Elkatatny & King Fahd University of Petroleum \& Minerals & 24 & 2021 \\
\hline
3 & New approach to evaluate the equivalent circulating density (ECD) using artificial intelligence techniques \cite{AbdelgawadKhaledZ.2019Nate} & Khaled Abdelgawad; Shirish Patil & King Fahd University of Petroleum and Minerals & 50 & 2019 \\
\hline
4 & The different member equivalent circulating density prediction model and drilling parameter optimization under narrow density window \cite{ZhaoWanchun2025Tdme} & Wanchun Zhao; Peihong Zhai & National Key Laboratory of Green Multi-Resource Collaborative Onshore Shale Oil Exploitation; Northeast Petroleum University & 0 & 2025 \\
\hline
5 & Bottom hole pressure prediction based on hybrid neural networks and Bayesian optimization \cite{ZhangCheng-Kai2023Bhpp} & Chengkai Zhang; Liang Han & China University of Petroleum & 17 & 2023 \\
\hline
6 & Intelligent Model for Predicting Downhole Vibrations Using Surface Drilling Data During Horizontal Drilling \cite{SaadeldinRamy2022IMfP} & Ramy Saadeldin; Abdulazeez Abdulraheem & King Fahd University of Petroleum and Minerals & 17 & 2022 \\
\hline
7 & Detecting downhole vibrations through drilling horizontal sections: machine learning study \cite{SaadeldinRamy2023Ddvt} & Ramy Saadeldin; Salaheldin Elkatatny & King Fahd University of Petroleum and Minerals & 14 & 2023 \\
\hline
8 & A Novel Hybrid Transfer Learning Method for Bottom Hole Pressure Prediction \cite{ZhangRui2023ANHT} & Rui Zhang; Chenxing Gong & China University of Petroleum; PetroChina Changqing Oilfield Company & 0 & 2023 \\
\hline
9 & An online hybrid prediction model for mud pit volume in the complex geological drilling process \cite{ZhouYang2021Aohp} & Yang Zhou; Takao Terano & China University of Geosciences; Chiba University of Commerce & 8 & 2021 \\
\hline
10 & Deep learning approach to prediction of drill-bit torque in directional drilling sliding mode: Energy saving \cite{CAOWanpeng2025Dlat} & Wanpeng Cao; Hamzeh Ghorbani & China Suntien Green Energy Corporation Limited; Islamic Azad University & 5 & 2025 \\
\hline
11 & Machine Learning-Based Trigger Detection of Drilling Events Based on Drilling Data \cite{ZhaoJie2017MLTD} & Jie Zhao; Sonny Johnston & Schlumberger & 34 & 2017 \\
\hline
12 & Downhole data correction for data-driven rate of penetration prediction modeling \cite{EncinasMauroA.2022Ddcf} & Mauro Encinas; Dan Sui & University of Stavanger; Northwestern Polytechnical University & 24 & 2022 \\
\hline
13 & Using Trees, Bagging, and Random Forests to Predict Rate of Penetration During Drilling \cite{HegdeChiranth2015UTBa} & Chiranth Hegde; Ken Gray & University of Texas & 109 & 2015 \\
\hline
\end{tabular}
\label{tab:reviewed-papers}
\end{table*}

\subsection{Data Extraction}
For each included paper, data was systematically extracted into a structured table. The table 
columns were as followed:
\begin{itemize}
\item Title: Provides reference point for the paper, ensuring that results are traceable and 
verifiable.
\item Surface Metrics Used: Addresses research question 1.1, identifying what surface metrics are 
collected.
\item Downhole Metrics Predicted: Addresses research question 1.2, identifying what downhole 
metrics are predicted.
\item ML Model Used: Addresses research question 2.1, identifying what techniques 
are currently used.
\end{itemize}

These columns were selected because they directly align with the research questions that guide 
the study. Model accuracy was not included because papers used different accuracy metrics (i.e., 
root mean squared error, mean absolute error, mean squared error, r-squared), predicted different units, 
and worked with datasets having different statistical characteristics (i.e., mean and 
standard deviation), making direct comparison of accuracy metrics across papers problematic.
The strengths and weaknesses of each method were not included for two reasons. First, assessing 
methodological strengths and weaknesses requires interpretive judgment, which risks reducing 
reproducibility. Second, papers tended to focus only on the strengths of their methods and the 
resulting accuracy, providing few weaknesses to interpret. Furthermore, accuracy and 
strengths/weaknesses were not included as columns because they do not directly answer any of 
the proposed research questions.

\section{Limitations}
This methodology has several limitations:
\begin{itemize}
\item This review is limited to only research published in English, which excludes non-English research.
\item Although multiple databases were used, the chosen databases do not cover every possible 
place that relevant studies may be indexed.
\item This review is limited from 2015 to 2025, potentially excluding earlier foundation work 
or future studies not yet published.
\end{itemize}

\section{Q1: What measurements are important to oil and gas drillers?}
Collected data related to drilling metrics are shown in Table~\ref{tab:surface-downhole}.

\begin{table*}[htbp]
\caption{Surface Metrics and Downhole Metrics of Reviewed Papers}
\centering
\begin{tabular}{|c|p{5.5cm}|p{6cm}|p{3cm}|}
\hline
\textbf{\#} & \textbf{Title} & \textbf{Surface Metrics} & \textbf{Downhole Metric} \\
\hline
1 & Explainable machine-learning-based prediction of equivalent circulating density using surface-based drilling data & ROP, WOB, WHO, RPM, T, Q, SPP, MW, TGO & ECD \\
\hline
2 & Machine Learning Models for Equivalent Circulating Density Prediction from Drilling Data & Q, ROP, RPM, SPP, WOB, T & ECD \\
\hline
3 & New approach to evaluate the equivalent circulating density (ECD) using artificial intelligence techniques & MW, DPP, ROP & ECD \\
\hline
4 & The different member equivalent circulating density prediction model and drilling parameter optimization under narrow density window & Q, RPM, ROP, WOB, SPP & ECD \\
\hline
5 & Bottom hole pressure prediction based on hybrid neural networks and Bayesian optimization & Depth, RPM, SPP, Q, TVD, Mud Volume, Sand Content, Back Pressure, Outlet Density, MW & BHP \\
\hline
6 & Intelligent Model for Predicting Downhole Vibrations Using Surface Drilling Data During Horizontal Drilling & Q, SPP, RPM, T, WOB, ROP & Downhole String Vibrations \\
\hline
7 & Detecting downhole vibrations through drilling horizontal sections: machine learning study & Q, SPP, RPM, T, WOB, ROP & Downhole String Vibrations \\
\hline
8 & A Novel Hybrid Transfer Learning Method for Bottom Hole Pressure Prediction & Depth, RPM, SPP, Q, Back Pressure, Outlet Flow, TVD, MW, Pool Volume, Outlet Density, Viscosity, Sand Content & BHP \\
\hline
9 & An online hybrid prediction model for mud pit volume in the complex geological drilling process & Depth, RPM, WOB, T, Q, SPP, MW, Conductivity, Temp & Mud Pit Volume \\
\hline
10 & Deep learning approach to prediction of drill-bit torque in directional drilling sliding mode: Energy saving & ROP, WOB, RPM & Drill Torque \\
\hline
11 & Machine Learning-Based Trigger Detection of Drilling Events Based on Drilling Data & WOB, RPM, Q, ROP, Block Position, Azm, Inc, DiffP & Drilling Event \\
\hline
12 & Downhole data correction for data-driven rate of penetration prediction modeling & Depth, Hookload, WOB, T, RPM, SPP & ROP Prediction \\
\hline
13 & Using Trees, Bagging, and Random Forests to Predict Rate of Penetration During Drilling & WOB, RPM, Q, ROP, Block Position, Azm, Inc, DiffP & ROP Prediction \\
\hline
\end{tabular}
\label{tab:surface-downhole}
\end{table*}

\subsection{Q1.1: What surface metrics are collected?}
\begin{figure}[htbp]
\centerline{\includegraphics[width=0.5\textwidth]{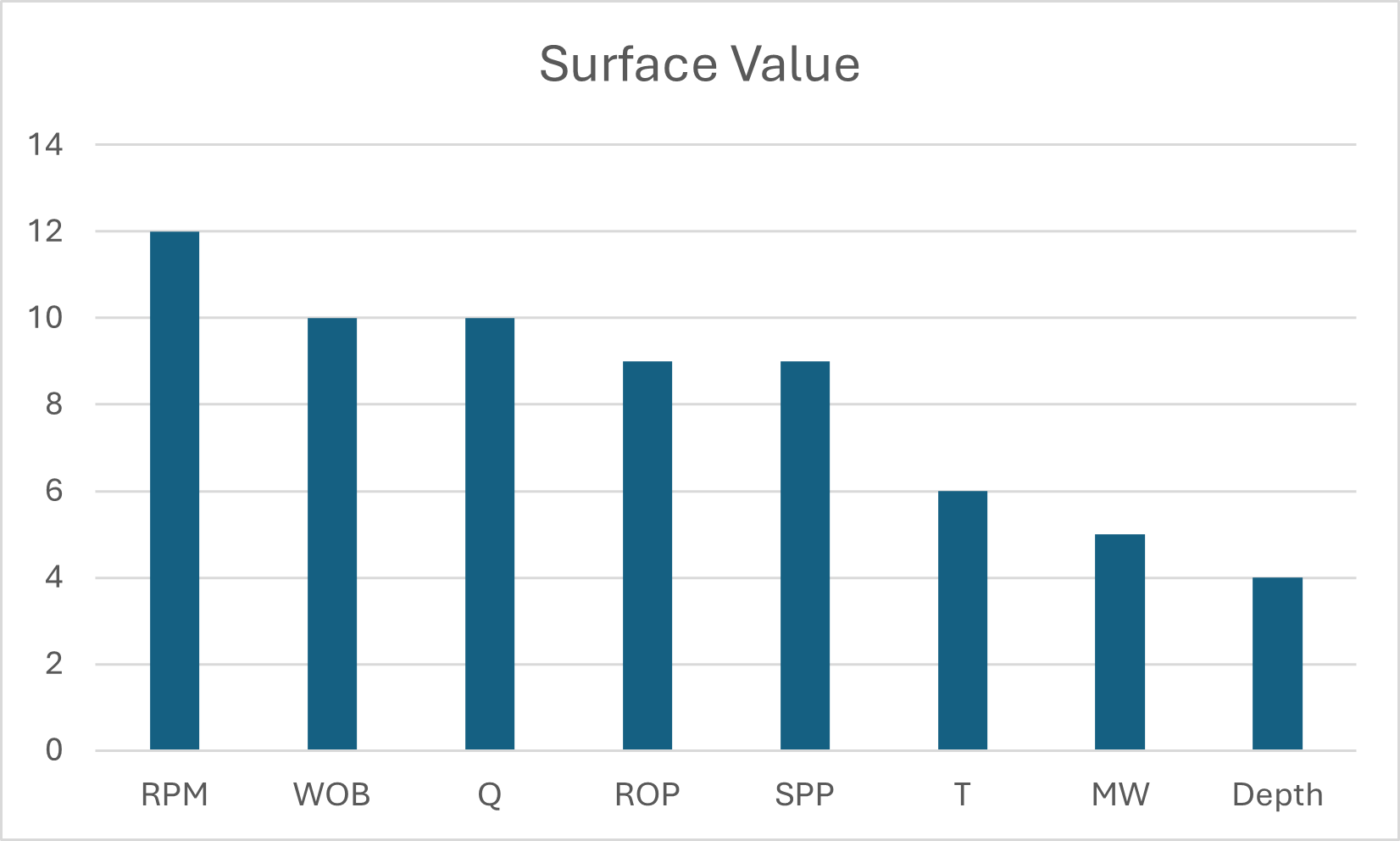}}
\caption{Bar chart illustrating the frequency of surface value frequencies across the reviewed papers.}
\label{surfaceChart}
\end{figure}

The collected data on surface metrics used are shown in Figure~\ref{surfaceChart}. Metrics that were used in two or fewer papers were considered outliers and discarded.

There were 8 different surface metrics collected, and they are defined as:
\begin{itemize}
\item Rotations Per Minute (RPM): The number of rotations the bit makes in a minute. 
This review does not differentiate between bit RPM and drillstring RPM \cite{drillingGlossary}.
\item Weight on Bit (WOB): The amount of downward force exerted on the drill bit \cite{drillingGlossary}. 
\item Flowrate (Q): Mud is the fluid that circulates through the drillstring and the wellbore \cite{drillingGlossary}. Flowrate refers to the 
volume of mud pumped through the drillstring per unit time. It’s usually expressed in gallons per minute 
or liters per minute \cite{iadc85}. 
\item Rate of Penetration (ROP): Drilling speed, usually expressed in distance per time unit 
(i.e., feet per hour, meters per second) \cite{drillingGlossary}.
\item Standpipe Pressure (SPP): The sum of all pressure losses that occur due to fluid friction in the 
wellbore, which is a sum of loss in annulus pressure, drillstring pressure, bottom hole assembly 
pressure, and drill bit pressure \cite{spp}.
\item Surface Torque (T): The amount of torque exerted on the drillstring \cite{drillingGlossary}.
\item Mud Weight (MW): Mud weight refers to the density of the mud \cite{drillingGlossary}. 
\item Depth: The depth of the wellbore, usually in meters or feet \cite{drillingGlossary}.
\end{itemize}
RPM was the most frequently used input to ML models, appearing in 12 of the 13 papers. WOB and Q were the next most used, each in 10 papers. ROP and SPP were used in 9 papers, 
T was used in 6, MW was used in 5, and depth was used in the fewest papers, appearing in just 4.

\subsection{Q1.2: What downhole metrics are important to predict?}
\begin{figure}[htbp]
\centerline{\includegraphics[width=0.5\textwidth]{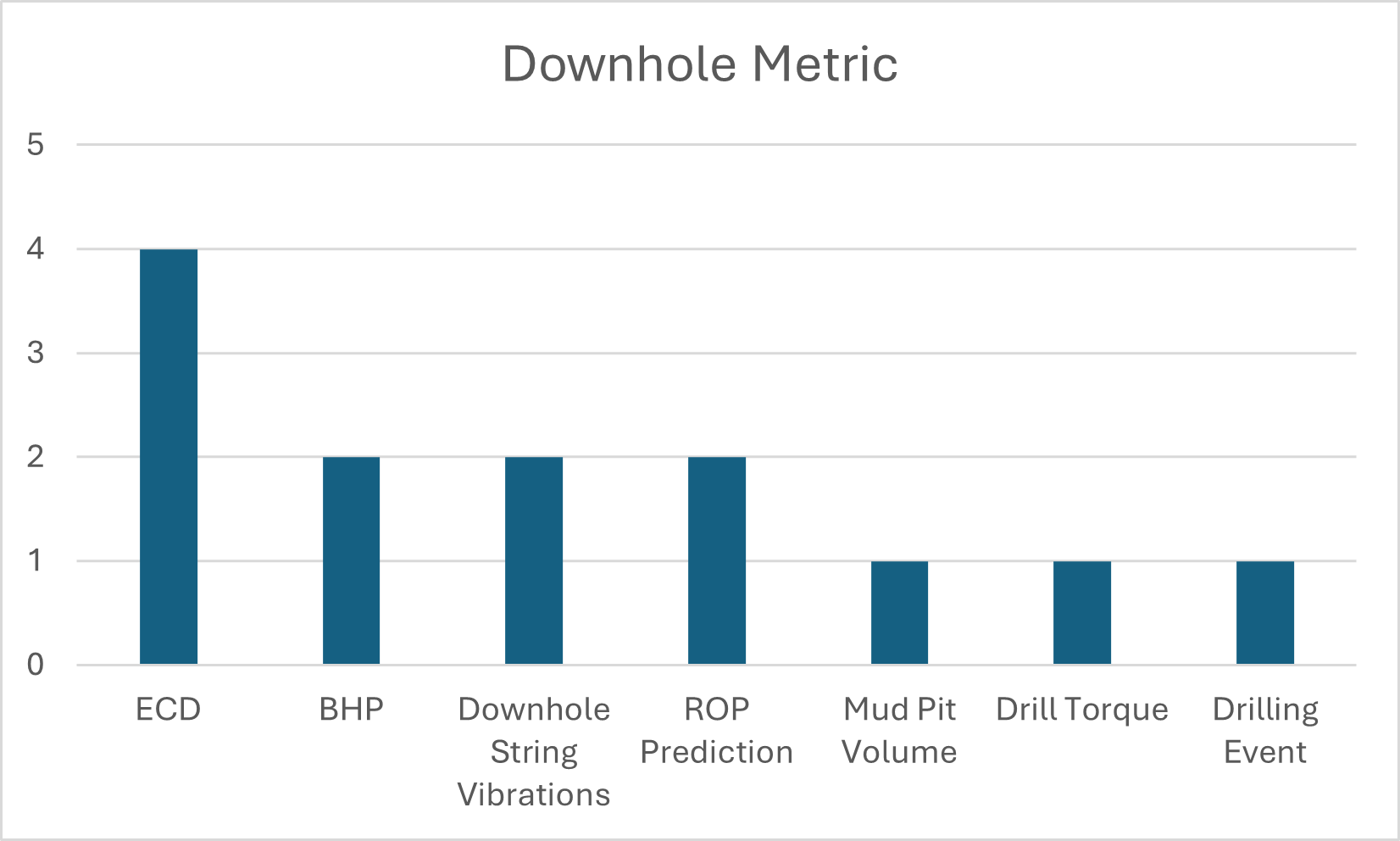}}
\caption{Bar chart illustrating the frequency of downhole metric frequencies across the reviewed papers.}
\label{downholeChart}
\end{figure}

The collected data on predicted downhole metrics are shown in Figure~\ref{downholeChart}.

There were 7 different predicted downhole metrics, defined as the following:
\begin{itemize}
\item Equivalent Circulating Density (ECD): Refers to the effective density exerted by the 
circulating mud against the formation being drilled. It considers the pressure drop in the 
annulus above the point being considered. It is important to avoid kicks and losses \cite{slbGlossary}. 
\item Bottom Hole Pressure (BHP): The summation of all pressures exerted on the bottom of the 
wellbore. Similar to ECD, it helps to avoid kicks in losses \cite{bhp}.  
\item Downhole String Vibrations: Refers to vibrations experienced by the drillstring. Too much 
vibration can lead to damage to the drillstring \cite{drillstringVibration}. 
\item ROP Prediction: Refers to predicting the downhole ROP.
\item Mud Pit Volume: A mud pit is a tank that holds mud on the rig. Mud pit volume refers to how 
much mud is in that tank. While not technically a downhole metric, the amount of mud currently downhole can be calculated using current mud pit volume \cite{slbGlossary}.
\item Drill Torque: Refers to predicting torque of the drillstring at the bottom of the wellbore, while T usually refers to 
torque at the surface \cite{torque}. 
\item Drilling Event: Any abnormal event in drilling is considered a drilling event, such as when part of the drillstring cannot be rotated or moved vertically (stuck pipe) \cite{slbGlossary} or an irregular movement caused by a buildup and then sudden drop of force on the drillstring due to it momentarily becoming stuck (stick-slip event) \cite{slbGlossary} \cite{10.2118/187512-MS}.
\end{itemize}

ECD was predicted in 4 papers, making it the most predicted downhole metric. This is probably due 
to how important it is to predict accurately and the difficulty in predicting it via rules-based 
algorithms \cite{10.56952/ARMA-2023-0722}. BHP, downhole string vibrations, and ROP prediction were 
all predicted in 2 papers. Finally, mud pit volume, drill torque, and drilling events were each 
predicted in just 1 paper. 

\section{Q2:  What methods are used to predict downhole metrics?}
Collected data related to prediction methodology are shown in Table~\ref{tab:ml-models}..

\begin{table*}[htbp]
\caption{ML Models Used in Reviewed Papers}
\centering
\begin{tabular}{|c|p{7cm}|p{6cm}|}
\hline
\textbf{\#} & \textbf{Title} & \textbf{ML Model} \\
\hline
1 & Explainable machine-learning-based prediction of equivalent circulating density using surface-based drilling data & XGBoost \\
\hline
2 & Machine Learning Models for Equivalent Circulating Density Prediction from Drilling Data & ANN, ANFIS \\
\hline
3 & New approach to evaluate the equivalent circulating density (ECD) using artificial intelligence techniques & ANN, ANFIS \\
\hline
4 & The different member equivalent circulating density prediction model and drilling parameter optimization under narrow density window & ANN \\
\hline
5 & Bottom hole pressure prediction based on hybrid neural networks and Bayesian optimization & RF, XGBoost, BPNN, CNN, GRU, CNN-GRU \\
\hline
6 & Intelligent Model for Predicting Downhole Vibrations Using Surface Drilling Data During Horizontal Drilling & ANN \\
\hline
7 & Detecting downhole vibrations through drilling horizontal sections: machine learning study & ANFIS, RBF, FN, SVM \\
\hline
8 & A Novel Hybrid Transfer Learning Method for Bottom Hole Pressure Prediction & LSTM-DANN \\
\hline
9 & An online hybrid prediction model for mud pit volume in the complex geological drilling process & SVR-BPNN-LSTM \\
\hline
10 & Deep learning approach to prediction of drill-bit torque in directional drilling sliding mode: Energy saving & DARN \\
\hline
11 & Machine Learning-Based Trigger Detection of Drilling Events Based on Drilling Data & SAX, hierarchical clustering \\
\hline
12 & Downhole data correction for data-driven rate of penetration prediction modeling & LSTM \\
\hline
13 & Using Trees, Bagging, and Random Forests to Predict Rate of Penetration During Drilling & Boosting, Pruned Tree, Random Forests, Decision Tree \\
\hline
\end{tabular}
\label{tab:ml-models}
\end{table*}

\subsection{Q2.1: What methods are currently used to predict downhole metrics from surface 
drilling data?}
The collected data on methods used to predict downhole metrics are shown in Table~\ref{tab:ml-method-frequency}. Note that several papers 
used multiple methods, thus more than 13 total methods are presented.

\begin{table}[htbp]
\caption{ML Methods and Their Frequency of Use}
\centering
\begin{tabular}{|p{6.5cm}|c|}
\hline
\textbf{ML Method Used} & \textbf{Times Used} \\
\hline
Artificial Neural Network (ANN) & 4 \\
\hline
Adaptive Neuro-Fuzzy Inference System (ANFIS) & 3 \\
\hline
Extreme Gradient Boosting (XGBoost) & 2 \\
\hline
Random Forest (RF) & 1 \\
\hline
Backpropagation Neural Network (BPNN) & 1 \\
\hline
Convolutional Neural Network (CNN) & 1 \\
\hline
Gated Recurrent Unit (GRU) & 1 \\
\hline
Convolutional Neural Network, Gated Recurrent Unit (CNN-GRU) & 1 \\
\hline
Radial Basis Function network (RBF) & 1 \\
\hline
Fuzzy Network (FN) & 1 \\
\hline
Support Vector Machine (SVM) & 1 \\
\hline
Long Short-Term Memory (LSTM) & 1 \\
\hline
Long Short-Term Memory, Domain-Adversarial Neural Network (LSTM-DANN) & 1 \\
\hline
Support Vector Regression, Backpropagation Neural Network, Long Short-Term Memory (SVR-BPNN-LSTM) & 1 \\
\hline
Deep Adversarial Neural Network (DARN) & 1 \\
\hline
Symbolic Aggregate Approximation (SAX) & 1 \\
\hline
Hierarchical Clustering & 1 \\
\hline
Boosting & 1 \\
\hline
Pruned Tree & 1 \\
\hline
Random Forests & 1 \\
\hline
Decision Tree & 1 \\
\hline
\end{tabular}
\label{tab:ml-method-frequency}
\end{table}

Both conventional ML algorithms (i.e., SVM, Random Forest) and more complex algorithms (i.e., LSTM, CNN) are shown to be used in the literature. Furthermore, a wide range of methods are shown to be used, as only 3 of the 21 methods were used in more than one paper. It is important 
to note that neural networks of some type (i.e., BPNN, FN) were used in 14 of 21 methods, showing 
them to be preferred in downhole metric prediction. 

\section{Q3: Can MAEFMs be used for oil and gas drilling?}

\subsection{Q3.1: What are MAEFMs?}
The term "Masked Autoencoder Foundation Model" can be broken down into four parts: model, 
foundation, autoencoder, and masked \cite{He_Chen_Xie_Li_Dollar_Girshick_2022}.

Model refers to MAEFMs being a type of ML model that takes inputs and produces an 
output based on learned weights.

The term "foundation" in MAEFMs refers to the shared base model that serves as a general representation layer for multiple downstream tasks. These models are 
pretrained on large-scale datasets to learn broad, domain-agnostic patterns within the data. The 
resulting representations can then be efficiently fine-tuned on smaller, task-specific datasets, 
enabling effective adaptation to diverse applications while reducing the need for extensive retraining 
\cite{Schneider_Meske_Kuss_2024}.

An autoencoder forms the foundational training mechanism for MAEFMs and is based on an encoder-decoder 
architecture. It consists of two main components: an encoder, which compresses high-dimensional input 
data into a lower-dimensional latent representation, and a decoder, which reconstructs the original 
input from this latent space. First introduced by Rumelhart et al. \cite{Rumelhart_Hinton_Williams_1986}, 
the autoencoder is trained to minimize the reconstruction loss between its input and output, thereby 
learning to preserve the most salient features of the data while discarding redundant or nonessential 
information.

\begin{figure}[htbp]
\centerline{\includegraphics[width=0.5\textwidth]{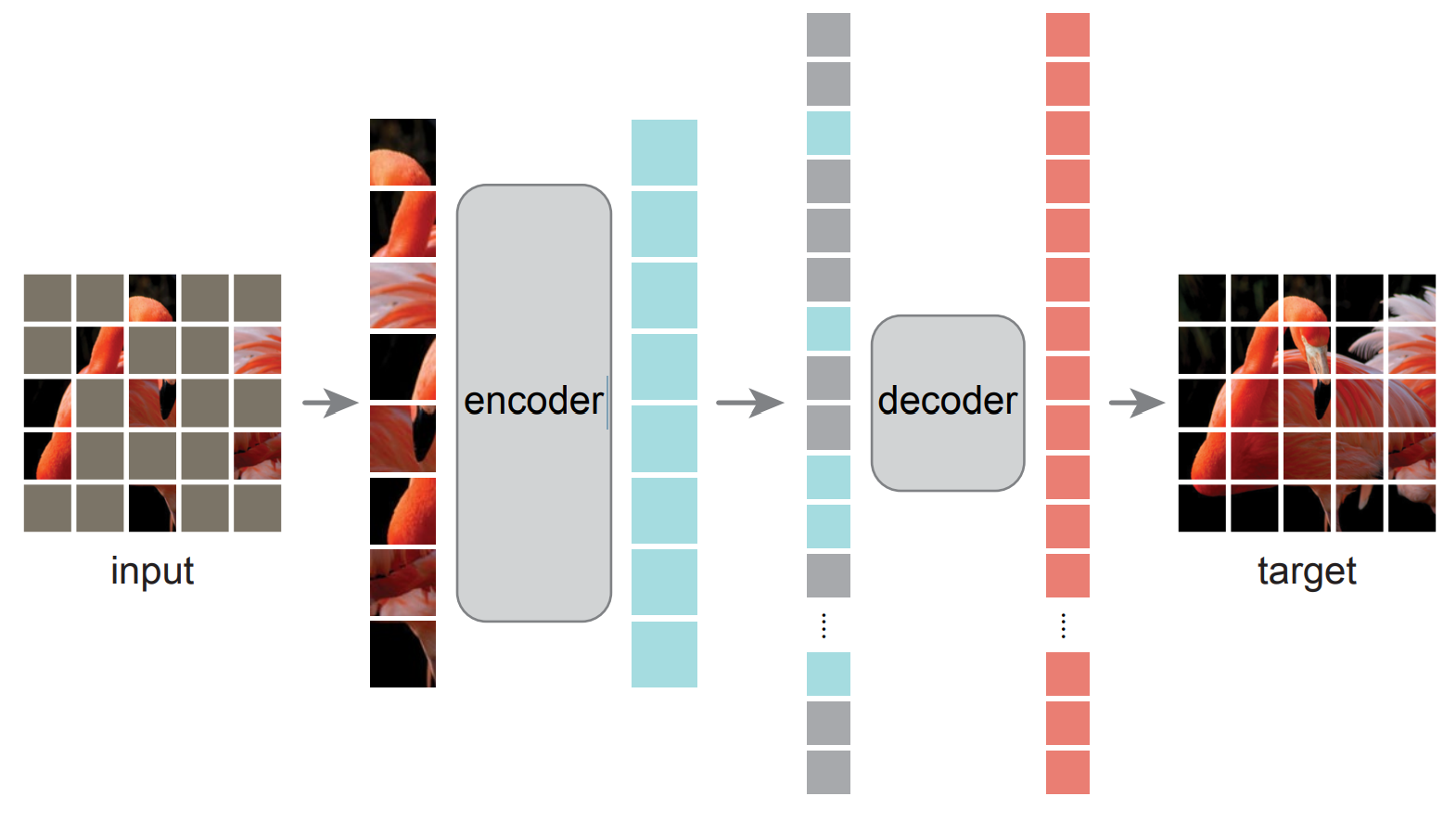}}
\caption{Masked autoencoder training architecture \cite{He_Chen_Xie_Li_Dollar_Girshick_2022}.}
\label{fig}
\end{figure}

The term "masked" refers to the deliberate omission of substantial portions of the input data during 
training of the autoencoder. By reconstructing the missing components, the model is encouraged to 
learn underlying structural and temporal patterns within the data, thereby capturing essential features 
more effectively \cite{Bachmann_Mizrahi_Atanov_Zamir_2022} \cite{He_Chen_Xie_Li_Dollar_Girshick_2022}. 

A critical aspect not conveyed by the term MAEFM is the use of task-specific headers. After pre-training 
the foundational model using the autoencoder architecture, the decoder is typically removed and the encoder 
weights are frozen. Task-specific layers are then appended to the frozen encoder and trained on the target 
task, preserving the pretrained foundation while enabling adaptation to multiple downstream tasks. This 
approach allows efficient utilization of smaller labeled datasets for training the task-specific adapters 
\cite{houlsby2019parameterefficienttransferlearningnlp}.

\subsection{Q3.2: Are MAEFMs suitable for time-series data such as surface drilling metrics?}
Li et al. evaluated masked autoencoders across five benchmark datasets (electricity transformer 
temperature, weather forecasting, exchange rate prediction, influenza case forecasting, and the 
UCR time-series archive \cite{UCRArchive2018}) and found that masked autoencoders consistently outperformed or matched other 
transformer-based architectures. This demonstrated their robustness and adaptability across 
diverse time-series domains, suggesting that the architecture can generalize to surface drilling 
metrics \cite{li2023timaeselfsupervisedmaskedtime}.

Tang et al. demonstrated that masked autoencoders are adept at handling multivariate time-series 
forecasting. Using three benchmark datasets (electricity transformer temperature, electricity consumption load, and weather forecasting), their study showed that multivariate masked autoencoders 
performed comparably to or better than LSTMs and other transformer models. This indicates that 
masked autoencoders can capture both temporal and cross-variable dependencies, a key requirement 
for modeling drilling processes where multiple correlated surface metrics (e.g., RPM, WOB, SPP, 
and flow rate) interact over time \cite{tang2022mtsmaemaskedautoencodersmultivariate}.

Li et al. and Tang et al. establish that masked autoencoder architectures are highly effective 
for time-series and multivariate temporal modeling. Since MAEFMs are trained using masked 
autoencoders, these findings indicate that MAEFMs should also exhibit strong performance on 
time-series and multivariate data. Consequently, MAEFMs appear not only suitable but potentially 
advantageous for oil and gas drilling applications, which inherently rely on complex, noisy, and 
multivariate time-series datasets \cite{li2023timaeselfsupervisedmaskedtime, 
tang2022mtsmaemaskedautoencodersmultivariate}.

\subsection{Q3.3: Are MAEFMs used in oil and gas drilling currently?}
As shown in Table~\ref{tab:ml-method-frequency}, MAEFMs, or more broadly, autoencoders of any type, are not currently used 
in the reviewed literature for downhole metric prediction. Among the 13 analyzed papers, none 
used autoencoder-based architectures. As previously stated, the dominant approach was instead 
neural networks such as ANNs, LSTMs, CNNs. All methods were focused on direct supervised learning for mapping surface metrics (e.g., RPM, WOB, SPP) to downhole targets (e.g., ECD, BHP), rather than the representation learning approaches that autoencoders and MAEFMs provide.

\section{Q4: What should the research direction for MAEFMs in the context of oil and gas drilling 
be?}

\subsection{Q4.1: What areas could MAEFMs be applied to in oil and gas drilling?}
MAEFMs can be applied to the prediction of any downhole metrics specified in Q1.2, which includes 
ECD, BHP, downhole string vibrations, ROP prediction, mud pit volume, drill torque, and drilling 
events. The inputs to such model would be the surface metrics specified in Q1.1, namely RPM, WOB, 
Q, ROP, SPP, and T.

Unlike traditional models that are each trained for a specific prediction task, a MAEFM can 
leverage large amounts of drilling data to learn generalizable representations of drilling 
dynamics and predict multiple downhole metrics simultaneously. MAEFMs enable a single generalized model to replace several specialized ones, reducing computational overhead, simplifying deployment, and improving maintainability and reliability on the rig.

\section{Discussion}

Drilling wellbores for oil and gas generates extensive unlabeled datasets comprised of surface metrics such as RPM, 
WOB, Q, ROP, SPP, T, MW, and depth. These eight surface metrics are currently utilized as inputs to ML models to 
calculate seven downhole parameters, namely ECD, BHP, downhole string vibrations, ROP prediction, mud pit volume, 
drill torque, and drilling events. Traditionally, these downhole metrics have been estimated using deterministic 
models; however, as shown, ML approaches are increasingly being employed. 

A wide variety of ML techniques have been applied to downhole metric prediction, with neural networks 
representing the most frequently used ML models. This diversity highlights the range of modeling strategies currently 
being explored and indicates that the introduction of novel ML paradigms remains feasible. Nonetheless, 
the dominance of neural network-based approaches underscores their suitability and preference for capturing 
the complex, nonlinear relationships inherent in drilling data.

MAEFMs, a neural network-based architecture, have been demonstrated to 
effectively model time-series data. Given their ability to capture temporal and multivariate dependencies, MAEFMs 
are well suited for predicting downhole metrics from surface measurements. The underlying masked autoencoder 
framework enables the model to learn meaningful latent representations even from partially observed or unlabeled 
sequences, which aligns closely with the characteristics of drilling datasets.

A key advantage of MAEFMs is their capacity to leverage large volumes of unlabeled drilling data to develop 
generalizable representations of drilling dynamics. These representations can then support the prediction of 
multiple downhole metrics simultaneously. By consolidating several task-specific models into a single generalized 
model, MAEFMs have the potential to reduce computational requirements, simplify deployment, and enhance 
maintainability and reliability in operational settings.

Overall, MAEFMs are well-positioned to offer a novel approach to downhole metric prediction. They perform 
effectively when provided with extensive unlabeled surface data alongside smaller labeled datasets containing 
both surface and downhole measurements. Consequently, MAEFMs can exploit the vast quantities of historical and 
real-time drilling data already generated in the industry. Furthermore, once pretrained on historical datasets, 
MAEFMs can be adapted to new wells and emerging drilling tasks, offering a flexible and scalable solution for 
advanced drilling analytics.

\section{Future Research Directions}
Future research should begin by evaluating whether MAEFMs can predict downhole metrics with accuracy comparable 
to that of existing ML models. Although current literature suggests that MAEFMs are capable of achieving 
this, empirical validation remains essential. Initial studies should focus on single-task models to establish a 
baseline for performance and reliability before progressing to multi-task models that can predict multiple 
downhole metrics simultaneously. Careful comparisons between MAEFMs and currently employed models, particularly 
neural networks, will be crucial not only in terms of predictive accuracy but also with regard to generalizability 
across different wells, formations, and drilling conditions. Additionally, understanding the amount of labeled 
data required for effective training of MAEFMs relative to other ML approaches will be an important 
consideration, as these models are expected to require substantially fewer labeled examples due to their ability 
to leverage large quantities of unlabeled data.

If MAEFMs are successfully applied to downhole metric prediction, their applicability may extend to other areas 
of oil and gas operations where representation learning is advantageous. For example, they could be employed in 
thermal modeling for oil sands extraction or other processes that are difficult to monitor in real time. More 
broadly, any task characterized by limited real-time measurements, a shared set of observable inputs, and 
challenges for traditional rule-based modeling could potentially benefit from MAEFMs. This positions them as a 
flexible and scalable tool for predictive analytics across a wide range of complex, data-rich operations in the 
oil and gas industry.

\section{Conclusion}
This study investigated the potential for MAEFMs 
to be applied in oil and gas drilling. Thirteen papers from the last 10 years were reviewed to 
identify commonly used surface metrics, predicted downhole metrics, and existing ML 
approaches. The review found that RPM, WOB, and Q 
were the most frequently collected surface metrics, while ECD 
and BHP were the most frequently predicted downhole metrics. The analysis 
also revealed that neural network-based approaches dominate the current literature, with methods such 
as ANNs, LSTMs, and CNNs appearing most frequently. Notably, no reviewed studies utilized 
autoencoder-based or foundation model architectures.

Given the demonstrated success of masked autoencoder models in other time-series domains, MAEFMs 
represent a technically feasible yet unexplored approach for drilling analytics. Their ability to 
learn generalized representations across multiple datasets and predict several downhole metrics 
simultaneously makes them a strong candidate for future research and development in drilling 
optimization. Future work should focus on pre-training MAEFMs on large-scale drilling datasets, 
evaluating their performance against established deep learning models, and exploring their 
ability to generalize across different wells and formations.

By highlighting both the current state of ML in drilling and the absence of foundation model 
research, this study identifies a clear opportunity for innovation at the intersection of 
subsurface engineering and advanced ML.

\section{References}
\renewcommand{\refname}{}
\bibliographystyle{IEEEtran}
\bibliography{LiteratureReview}

\end{document}